\documentclass[letterpaper, 10pt, conference]{ieeeconf} 
\makeatletter
\let\NAT@parse\undefined
\makeatother

\usepackage{dblfloatfix}
\usepackage[numbers, sectionbib, sort]{natbib}
\usepackage{bm}
\usepackage{gensymb}
\usepackage{xcolor}
\usepackage{graphicx}
\usepackage{amsmath}
\usepackage{amssymb}
\usepackage{subcaption}
\usepackage{amsfonts}
\usepackage{multicol}
\usepackage{siunitx}
\usepackage{booktabs}
\usepackage{makecell}
\usepackage{multirow}
\usepackage{upgreek}
\usepackage[font=small]{caption}
\usepackage[export]{adjustbox}
\usepackage{tikz}
\usepackage{tabularx}
\usepackage{sidecap} \sidecaptionvpos{figure}{c}
\usepackage[hidelinks]{hyperref}
\usepackage[nameinlink, capitalize]{cleveref}
\usepackage[printonlyused,withpage,nolist,nohyperlinks]{acronym}
\usepackage{float} 
\usepackage{afterpage}  
\usepackage{placeins} 
\usepackage[T1]{fontenc}
\usepackage{balance}

\usepackage{cuted}
\usepackage[ruled,linesnumbered]{algorithm2e}
\SetNlSty{}{}{:}     
\SetKwInput{KwReq}{Require}

\Crefname{section}{Sec.}{Sec.}
\Crefname{equation}{Eq.}{Eq.}
\newcommand\etal{\emph{et al.}}

\begin{document}
\title{\LARGE \bf
Designing Privacy-Preserving Visual Perception for Robot Navigation \\ Based on User Privacy Preferences
}

\author{
Xuying Huang \and Sicong Pan \and Delphine Reinhardt \and Maren Bennewitz}

\maketitle

\begin{abstract}
Visual navigation is a fundamental capability of mobile service robots, yet the onboard cameras required for such navigation can capture privacy-sensitive information and raise user privacy concerns. 
Existing approaches to privacy-preserving navigation-oriented visual perception have largely been driven by technical considerations, with limited grounding in user privacy preferences. 
In this work, we propose a user-centered approach to designing privacy-preserving visual perception for robot navigation. 
To investigate how user privacy preferences can inform such design, we conducted two user studies. 
The results show that users prefer privacy-preserving visual abstractions and capture-time low-resolution preservation mechanisms: their preferred RGB resolution depends both on the desired privacy level and robot proximity during navigation. 
Based on these findings, we further derive a user-configurable distance-to-resolution privacy policy for privacy-preserving robot visual navigation.
\end{abstract}

{%
\renewcommand\thefootnote{}% 
\footnotetext{%
X. Huang, S. Pan, and M. Bennewitz are with the Humanoid Robots Lab, University of Bonn, the Lamarr Institute for Machine
Learning and Artificial Intelligence and the Center for Robotics, Bonn, Germany. 
D. Reinhardt is affiliated to the Computer Security and Privacy, University of Goettingen, Goettingen, Germany.
This work has been partially funded by the
German Federal Ministry of Research, Technology and Space~(BMFTR) under grant No. 16KIS1949 and the Robotics Institute Germany (RIG).}%}\
}%
\setcounter{footnote}{0}

\section{INTRODUCTION} 
Visual navigation is a core capability of mobile service robots operating in human-centered environments such as homes, offices, healthcare facilities, and classrooms~\citep{Bonin2008jirs, Araceli2019jirs, privatar2023}. 
To navigate effectively, such robots often rely on onboard cameras that continuously observe the surrounding environment. 
However, as mobile service robots traverse these spaces, their visual ability raises user privacy concerns~\citep{reinhardt2021pmc}, since a robot may inadvertently capture privacy-sensitive content, such as human faces, computer screens, identity documents, or personal belongings~\citep{kqiku2026percum}.

To mitigate such privacy risks, prior works have explored several ways to preserve privacy, including sanitization~\citep{Yang2025appliedsciences}, replacing raw RGB input with privacy-friendly representations~\citep{pietrantoni2023cvpr}, and ultra-low-resolution sensing from the source~\citep{huang2_2025arxiv}. 
However, the underlying design choices are typically motivated from a technical perspective, without relying on a thorough evaluation of user privacy preferences. Existing studies on Human-Robot Interaction (HRI) have shown that user privacy preferences toward mobile robots are nuanced and context-dependent~\citep{Rueben2017hri, Dietrich2023frai}. Nevertheless, these user-centered insights have rarely been translated into visual perception design for robot navigation.
 
In this work, we propose a user-centered approach to designing privacy-preserving visual perception for robot navigation. 
Rather than treating privacy in robot navigation solely as a technical property of visual representations, we aim to ground the design of robot visual perception in user privacy preferences. 
This raises a central question: How can user privacy preferences translate into the design of privacy-preserving visual perception for robot navigation?

\begin{figure}[!t] 
  \centering
  \includegraphics[width=1.0\columnwidth]{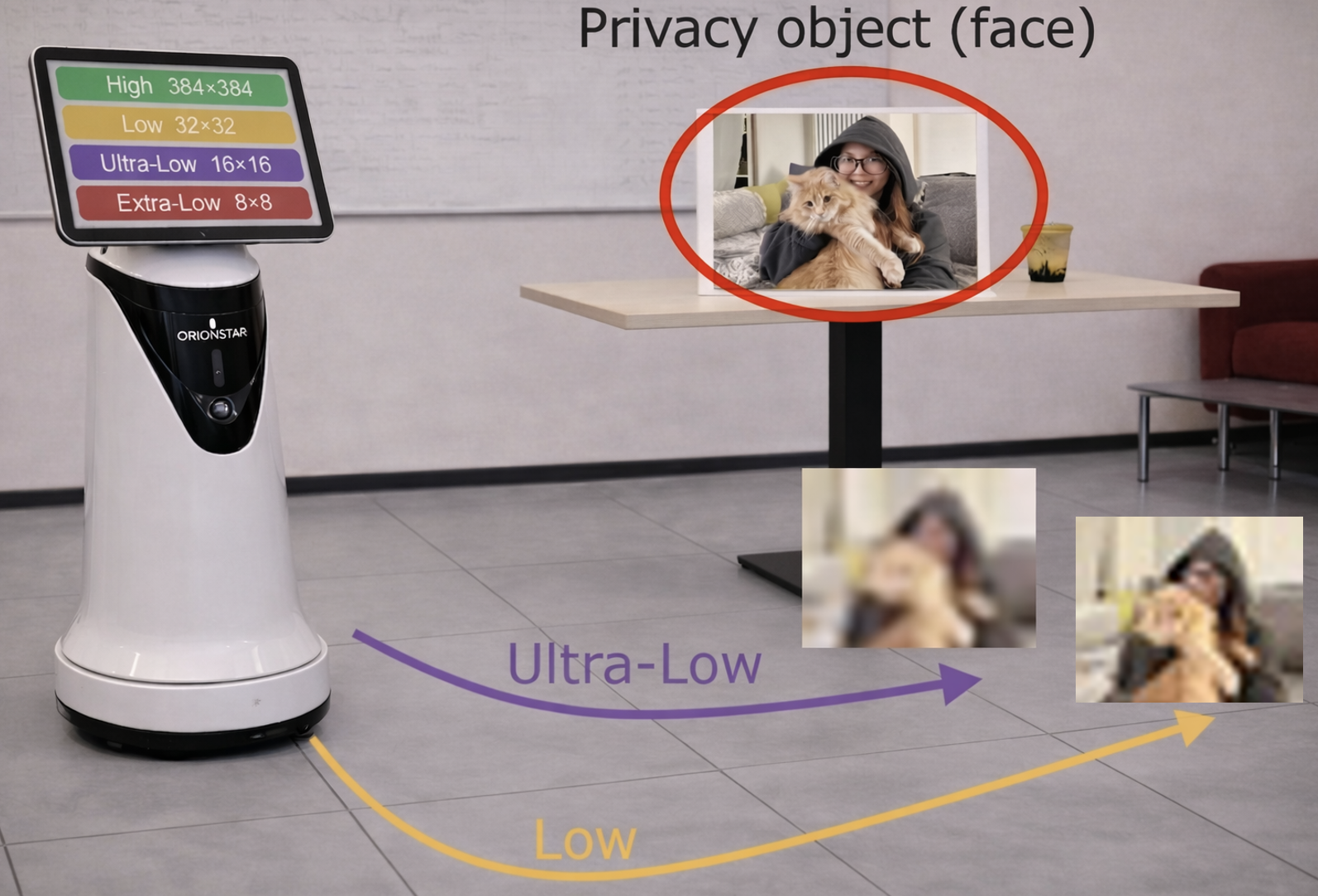}
  \caption{
  Illustration of a mobile service robot navigating in a private indoor environment.
  The privacy-sensitive object~(human face) is highlighted by the red circle. 
  Through a user-configurable multi-resolution RGB perception interface, different RGB resolutions can be selected by the users according to their privacy needs. 
  The example visualizations show how lowering resolution reduces the recognizability of privacy-sensitive content as the robot approaches the privacy object. 
  Based on our two user studies, this figure illustrates our central design implication: privacy-preserving robot navigation can be implemented by a user-configurable distance-to-resolution policy, adapting RGB resolution to both user privacy preferences and robot proximity.
  }
  \label{fig:teaser}
  \vspace{-0.3cm}
\end{figure}

To answer this question, we conducted two user studies. 
The first study investigates which visual inputs users perceive as privacy-preserving. 
The second study examines whether user privacy preferences change during robot navigation in privacy-sensitive scenarios. 
To the best of our knowledge, we are the first to consider user privacy preferences as a basis for visual perception configurations for robot navigation.
Together, these studies lead to a central design implication for robot navigation, illustrated in Fig.~\ref{fig:teaser}.

The primary contributions of our work are threefold:
\begin{itemize}
    \item \textbf{Study 1:} We characterize user privacy preferences over visual inputs, across both visual modalities and RGB resolution levels.
    \item \textbf{Study 2:} We show that user privacy preferences for RGB resolution vary during robot navigation, changing with robot proximity to privacy-sensitive objects.
    \item \textbf{Design implication:} We derive a user-configurable distance-to-resolution privacy policy and show that it can be readily instantiated in an existing robot navigation system with multi-resolution RGB inputs, providing a practical way to operationalize user privacy preferences in robot navigation.
\end{itemize}

\section{RELATED WORK} 
User privacy in mobile robots has attracted growing attention, as they become increasingly prevalent in everyday environments~\citep{rueben2016wr}. 
Prior work has characterized privacy in robotic systems as a multi-dimensional construct, commonly including informational, physical, psychological, and social privacy~\citep{reinhardt2021pmc}. 
Among these dimensions, informational privacy is particularly relevant to robot navigation, since onboard cameras can capture and process sensitive visual information during operation~\citep{taras2023arXiv}. 

To mitigate such risks, prior works have explored several privacy-preserving strategies for robot visual navigation. 
One line of work sanitizes robot visual streams by concealing~\citep{Choi2025arxiv} or filtering~\citep{Yang2025appliedsciences} privacy-sensitive content, thereby limiting the exposure of identifiable visual details during robot operation.
Another line of work replaces raw RGB imagery with more privacy-preserving representations. For example, SegLoc~\citep{pietrantoni2023cvpr} leverages privacy-preserving semantic segmentation for visual localization. MOSAIC~\citep{liu2025mosaic} generates consistent, privacy-preserving scene reconstructions from multiple depth views.
Recent works have investigated data minimization directly through Ultra-Low-Resolution~(ULR) RGB acquisition during robot navigation such as Huang~\etal~\citep{huang2_2025arxiv} who proposed a privacy-preserving semantic segmentation pipeline that uses $16 \times 16$ as ULR inputs for semantic object-goal navigation.
While low RGB resolutions can enhance privacy preservation, they may degrade semantic segmentation quality and, consequently, downstream navigation performance~\citep{Chaplotneurips, Florarxiv}, motivating adaptive rather than uniformly low resolution settings.

However, these approaches address privacy from a technical perspective. They operationalize privacy through visual abstractions, privacy-preserving modalities or sensing constraints, without considering user privacy preferences. 
In contrast, our work leverages user privacy preferences derived from two user studies to inform the design of visual perception for robot navigation.

\begin{figure}[t]
    \centering
    \begin{subfigure}[t]{0.30\linewidth}
        \centering
        \includegraphics[width=\linewidth]{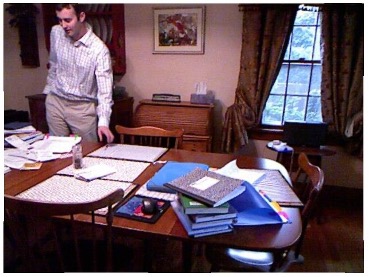}
        \caption{}
        \label{fig:rgb}
    \end{subfigure}\hfill
    \begin{subfigure}[t]{0.30\linewidth}
        \centering
        \includegraphics[width=\linewidth]{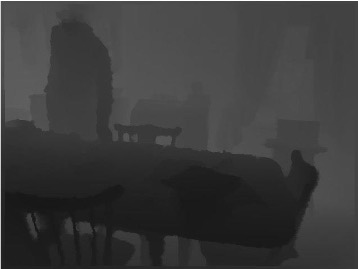}
        \caption{}
        \label{fig:depth}
    \end{subfigure}\hfill
    \begin{subfigure}[t]{0.30\linewidth}
        \centering
        \includegraphics[width=\linewidth]{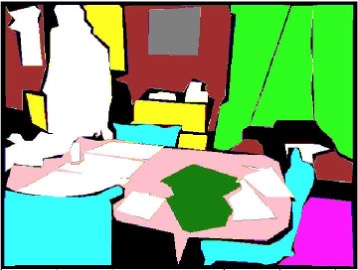}
        \caption{}
        \label{fig:semantic}
    \end{subfigure}
    \caption{Modality-comparison examples: (\subref{fig:rgb}) RGB image, (\subref{fig:depth}) depth image, and (\subref{fig:semantic}) semantic segmentation image.}
    \label{fig:modality}
    \vspace{-0.5cm}
\end{figure}

\begin{figure}[t]
    \centering
    \begin{subfigure}[t]{0.48\linewidth}
        \centering
        \includegraphics[width=\linewidth]{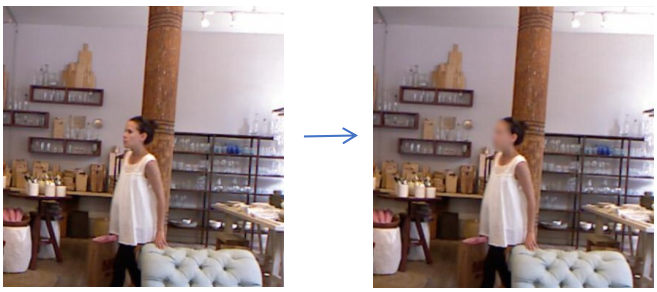}
        \caption{}
        \label{fig:post}
    \end{subfigure}\hfill
    \begin{subfigure}[t]{0.48\linewidth}
        \centering
        \includegraphics[width=\linewidth]{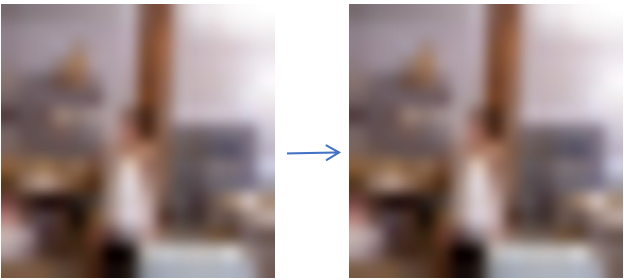}
        \caption{}
        \label{fig:capture}
    \end{subfigure}

    \caption{Examples to compare two privacy-preserving image-processing strategies: (\subref{fig:post}) post-processing of a high-resolution image and (\subref{fig:capture}) capture-time low-resolution sensing from the source.}
    \label{fig:processing}
    \vspace{-0.3cm}
\end{figure}

\begin{figure}[t]
    \centering

    \begin{subfigure}[t]{0.30\linewidth}
        \centering
        \includegraphics[width=\linewidth]{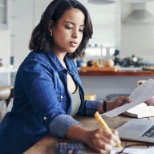}
        \caption{}
        \label{fig:384}
    \end{subfigure}\hfill
    \begin{subfigure}[t]{0.30\linewidth}
        \centering
        \includegraphics[width=\linewidth]{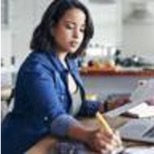}
        \caption{}
        \label{fig:96}
    \end{subfigure}\hfill
    \begin{subfigure}[t]{0.30\linewidth}
        \centering
        \includegraphics[width=\linewidth]{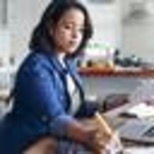}
        \caption{}
        \label{fig:64}
    \end{subfigure}

    \vspace{0.2em}

    \begin{subfigure}[t]{0.30\linewidth}
        \centering
        \includegraphics[width=\linewidth]{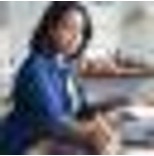}
        \caption{}
        \label{fig:32}
    \end{subfigure}\hfill
    \begin{subfigure}[t]{0.30\linewidth}
        \centering
        \includegraphics[width=\linewidth]{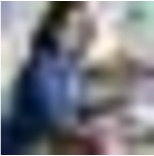}
        \caption{}
        \label{fig:16}
    \end{subfigure}\hfill
    \begin{subfigure}[t]{0.30\linewidth}
        \centering
        \includegraphics[width=\linewidth]{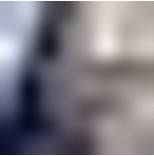}
        \caption{}
        \label{fig:8}
    \end{subfigure}
    \caption{Face example at six RGB resolutions: (\subref{fig:384}) $384 \times 384$, (\subref{fig:96}) $96 \times 96$, (\subref{fig:64}) $64 \times 64$, (\subref{fig:32}) $32 \times 32$, (\subref{fig:16}) $16 \times 16$, and (\subref{fig:8}) $8 \times 8$.}
    \label{fig:resolutions}
    \vspace{-1.0cm}
\end{figure}

\section{Study 1: User Preferences for Privacy-Preserving Visual Inputs}\label{sec:study_1}

\subsection{Study Design} 
We conducted an online questionnaire-based user study to evaluate user privacy preferences across four primary blocks. 
First, we assessed users’ comfort with robots having full visual access to private scenes and information. 
Second, we examined how different image modalities are perceived in terms of privacy intrusion and privacy preservation by presenting different visual representations. 
For this purpose, participants were shown matched examples of sensitive information in RGB, depth, and semantic segmentation as shown in Fig.~\ref{fig:modality}. 
These modality-comparison examples were derived from the NYU Depth V2 dataset~\citep{silberman2012eccv}. 
Third, participants were asked to indicate which image-processing strategy they preferred, i.e., post-processing of a high-resolution image or capture-time low-resolution sensing from the source~(as illustrated in Fig.~\ref{fig:processing}).
Finally, to assess RGB~resolution thresholds for privacy preservation, users were presented with a series of progressively downsampled RGB images and asked to select both minimal and maximum resolution they would find suitable for privacy preservation.
As shown in Fig.~\ref{fig:resolutions}, the candidate images spanned six resolution levels: $384\times384$, $96\times96$, $64\times64$, $32\times32$, $16\times16$, and $8\times8$. 
These RGB examples were generated using GPT-4o.

Participants evaluated the privacy preferences through mixed subjective measures. 
Specifically, the first block used one binary response and 5-point Likert-scale~(1 = very uncomfortable, 5 = very comfortable). 
The second block assessed image modalities through multi-select categorical judgments. 
The third block was recorded as a forced-choice preference and 5-point Likert-scale. 
The fourth block measured RGB privacy thresholds as ordinal selections. 

\subsection{Study Platform} The questionnaire was conducted via FreeOnlineSurveys platform.\footnote{\url{https://app.freeonlinesurveys.com/}}
Participants were recruited through public online channels, including social media and mailing lists.

\subsection{Participants} A total of N=62 individuals participated in this study. They were on average 27.2 years old~(SD = 5.6 years).

\begin{figure}[!t]
  \centering

  \begin{subfigure}[t]{\columnwidth}
    \centering
    \includegraphics[width=\linewidth]{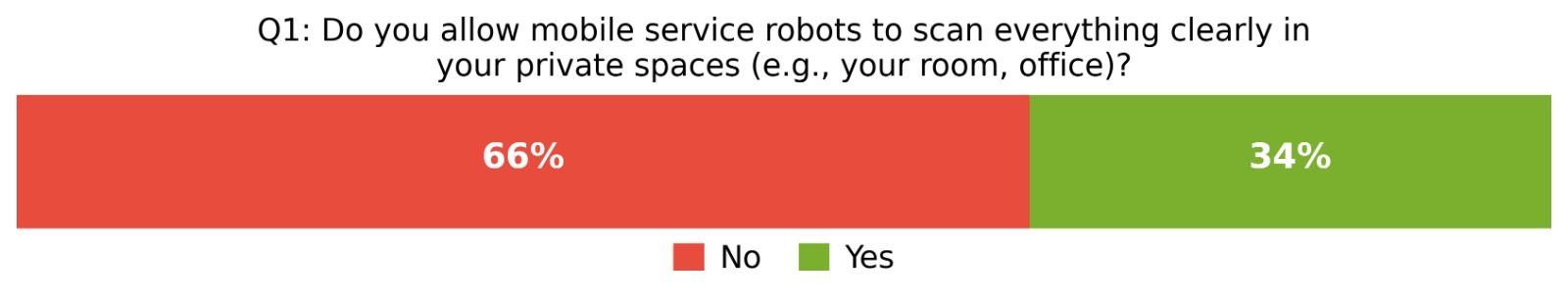}
    \caption{}
    \label{fig:allow}
  \end{subfigure}

  \vspace{0.0em}

  \begin{subfigure}[t]{\columnwidth}
    \centering
    \includegraphics[width=\linewidth]{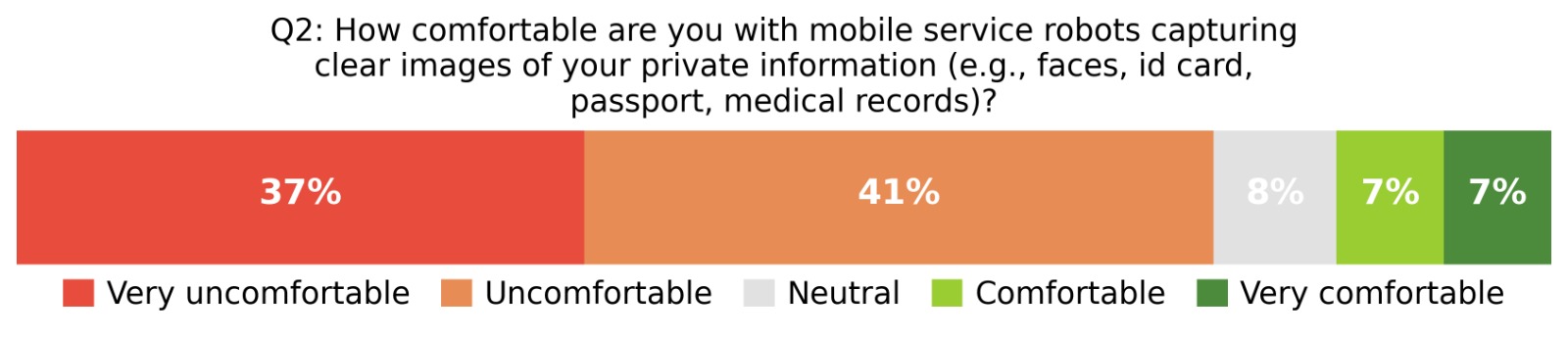}
    \caption{}
    \label{fig:comfort}
  \end{subfigure}

  \caption{
  Responses for first block regarding (\subref{fig:allow}) full robot vision and (\subref{fig:comfort}) clear capture of private information
  }
  \label{fig:full_vision}
  \vspace{-0.3cm}
\end{figure}

\begin{figure}[!t]
  \centering
  \begin{subfigure}[t]{0.48\columnwidth}
    \centering
    \includegraphics[width=\linewidth]{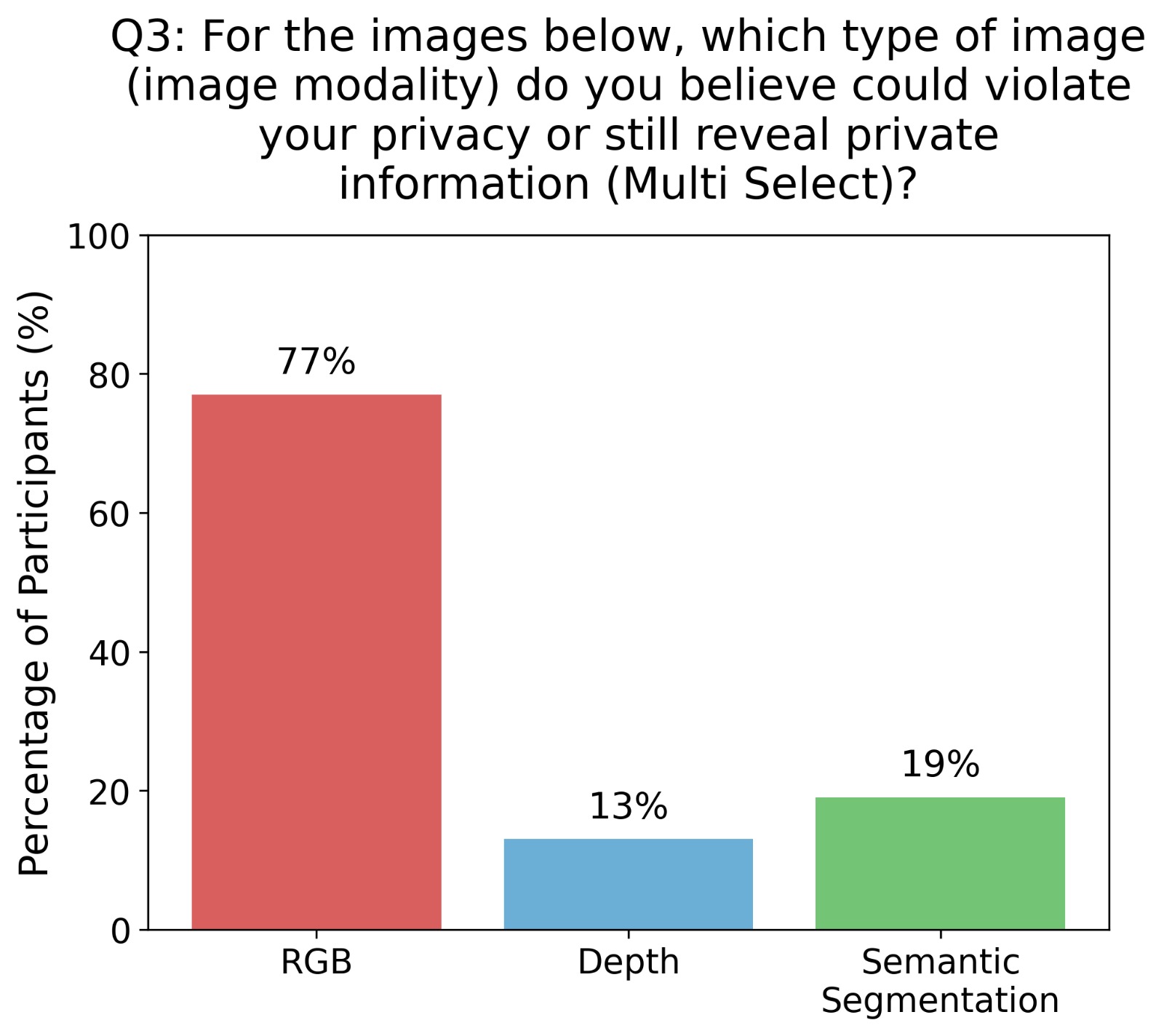}
    \caption{}
    \label{fig:modality_intrusive}
  \end{subfigure}\hfill
  \begin{subfigure}[t]{0.48\columnwidth}
    \centering
    \includegraphics[width=\linewidth]{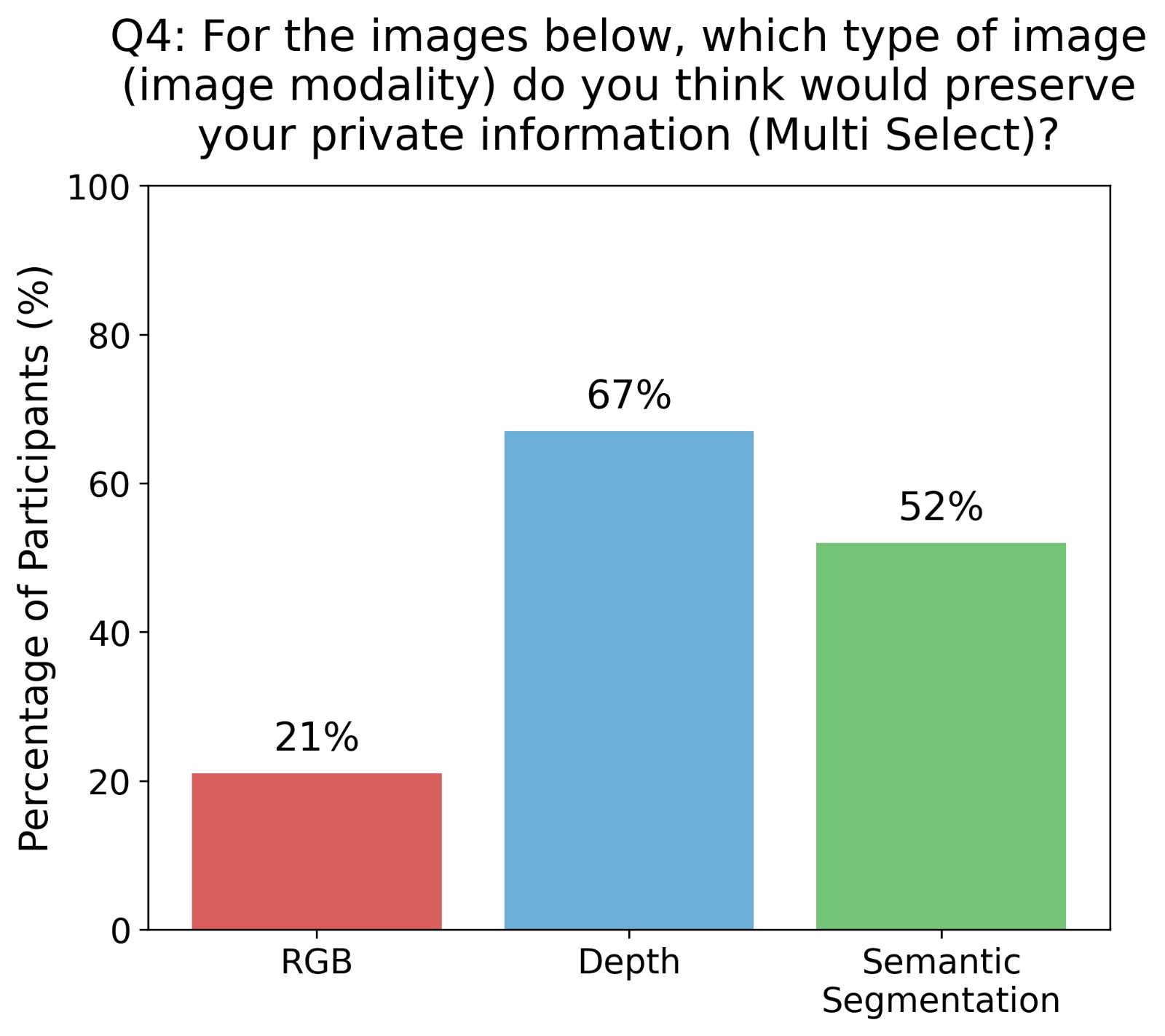}
    \caption{}
    \label{fig:modality_preserve}
  \end{subfigure}

  \caption{
  Responses for first block regarding perceived modalities that (\subref{fig:modality_intrusive}) violate privacy and (\subref{fig:modality_preserve}) preserve private information.
  }
  \label{fig:modality_result}
  \vspace{-0.3cm}
\end{figure}

\subsection{Study Results} 
\subsubsection{User Attitudes Towards Full Visibility} 
As shown in Fig.~\ref{fig:allow}, participants generally rejected full robot vision in private spaces: 66\% did not allow robots to clearly scan everything in their private spaces.
This is confirmed by the question focused specifically on privacy-sensitive information. 
As shown in Fig.~\ref{fig:comfort}, 
the majority of participants reported discomfort with robots capturing clear images of private information, with a total of 78\% being overall uncomfortable.
These results indicate that a majority of participants are uncomfortable with robots acquiring clear visual data in privacy-relevant contexts, especially when the captured content contains explicit private information.

\subsubsection{User Privacy Perception Across Visual Modalities} 
As presented in Fig.~\ref{fig:modality_result}, participants clearly differentiated the privacy perception of different visual modalities.  
RGB was most frequently perceived as privacy-intrusive, with 77\% of participants identifying it as capable of violating privacy or revealing private information, as illustrated in Fig.~\ref{fig:modality_intrusive}. 
In contrast, as illustrated in Fig.~\ref{fig:modality_preserve}, depth and semantic segmentation were more often perceived as privacy-preserving, reducing personally revealing details.
These findings suggest that participants associated raw RGB imagery with threats to privacy, whereas more abstract visual representations, such as depth and semantic segmentation, were perceived as more privacy-friendly.

\subsubsection{User Preferences for Privacy-Preserving Image Processing} 
\begin{figure}[!t]
  \centering

  \begin{subfigure}[t]{\columnwidth}
    \centering
    \includegraphics[width=\linewidth]{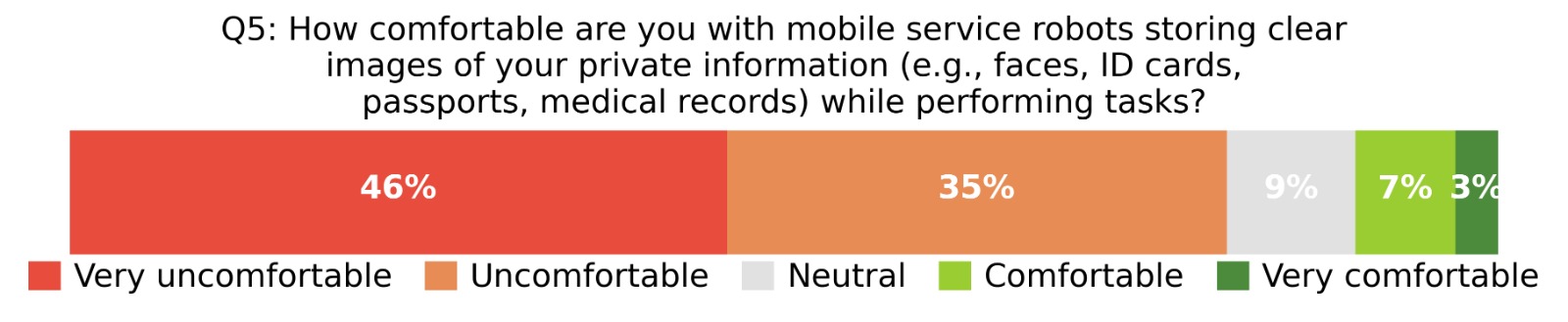}
    \caption{}
    \label{fig:storing}
  \end{subfigure}

  \vspace{0.0em}

  \begin{subfigure}[t]{\columnwidth}
    \centering
    \includegraphics[width=\linewidth]{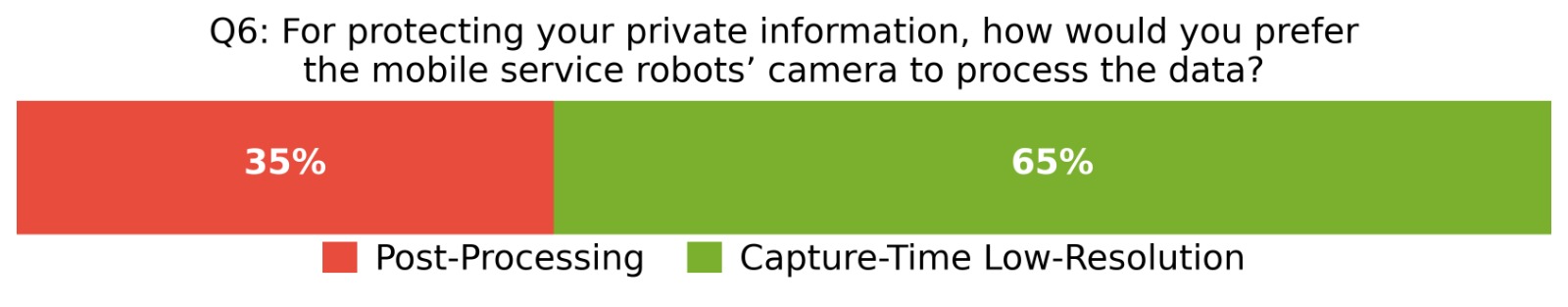}
    \caption{}
    \label{fig:strategy}
  \end{subfigure}

  \caption{
  Responses for third block regarding (\subref{fig:storing}) storage of clear private images and (\subref{fig:strategy}) preferred privacy-preserving processing strategy.
  }
  \label{fig:image_process}
  \vspace{-0.3cm}
\end{figure}

As shown in Fig.~\ref{fig:storing}, most participants were uncomfortable with robots storing clear images of private information during task execution, indicating limited acceptance of retaining privacy-sensitive visual records.
Fig.~\ref{fig:strategy} further shows a preference for capture-time low-resolution processing over post-processing as a privacy-preserving strategy. 
This suggests that participants prefer privacy preservation to be applied before sensitive visual details are captured, rather than after image acquisition. 
Taken together, these findings indicate that participants are not only uncomfortable with storing clear private images, but also prefer privacy-preserving approaches that intervene as early as possible in the visual sensing process.

\begin{figure}[!t] 
  \centering
  \includegraphics[width=\columnwidth]{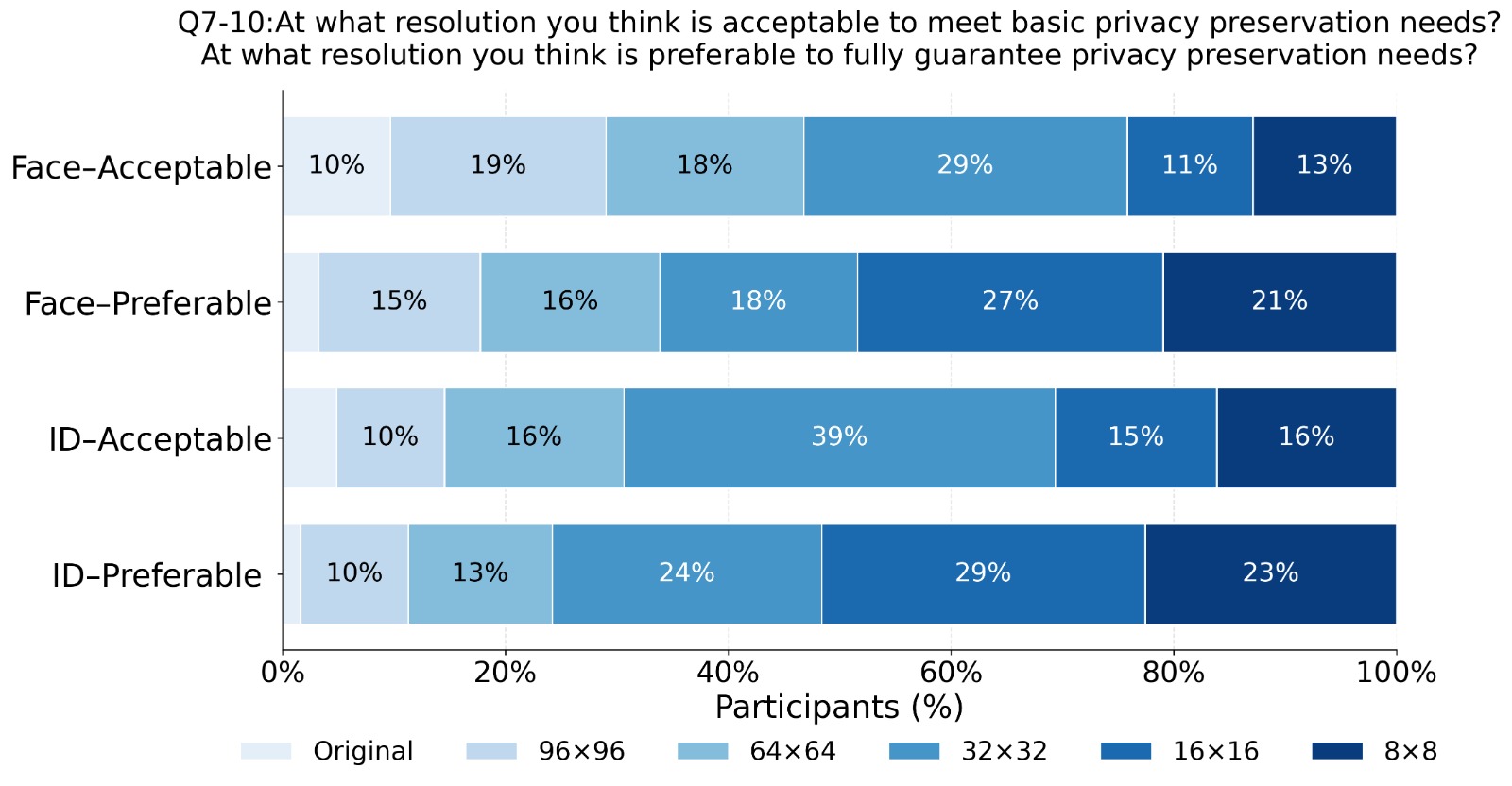}
  \caption{Responses for the fourth block, which examined users’ perceived RGB resolution thresholds for privacy preservation under low-resolution RGB images.  
}
  \label{fig:threshold}
  \vspace{-0.3cm}
\end{figure}

\subsubsection{User Preferences on RGB Resolution Thresholds for Privacy Preservation} Fig.~\ref{fig:threshold} shows that participants differentiated between acceptable and preferable RGB resolution thresholds for privacy preservation. 
For both face and ID card, responses in the acceptable condition concentrated at resolutions of $\leq 32 \times 32$, suggesting that participants generally regarded this range as sufficient for achieving a basic level of privacy preservation. 
However, when asked which resolution would preferably ensure privacy, responses shifted toward stronger downsampling. 
For both content types, preferences were more concentrated at $\leq 16 \times 16$, indicating a tendency to favor lower-resolution visual input under more privacy-preserving preferences. 
These results suggest that users hold different resolution-threshold preferences depending on the privacy preservation desired level.

\subsubsection{Summary}
Our results confirm that participants were sensitive to privacy risks in robot visual perception. 
They perceived depth and semantic segmentation as more privacy-preserving than raw original~RGB. 
For privacy-sensitive RGB input, they preferred low-resolution sensing at capture time over post-processing. 
The results further show that preferred RGB resolution thresholds depend on the desired privacy level: while $\leq 32 \times 32$ was commonly regarded as acceptable for basic privacy preservation, $\leq 16 \times 16$ was more often selected under the more privacy-preserving preference requirement. 

\begin{figure}[!t] 
  \centering
  \includegraphics[width=1.0\columnwidth]{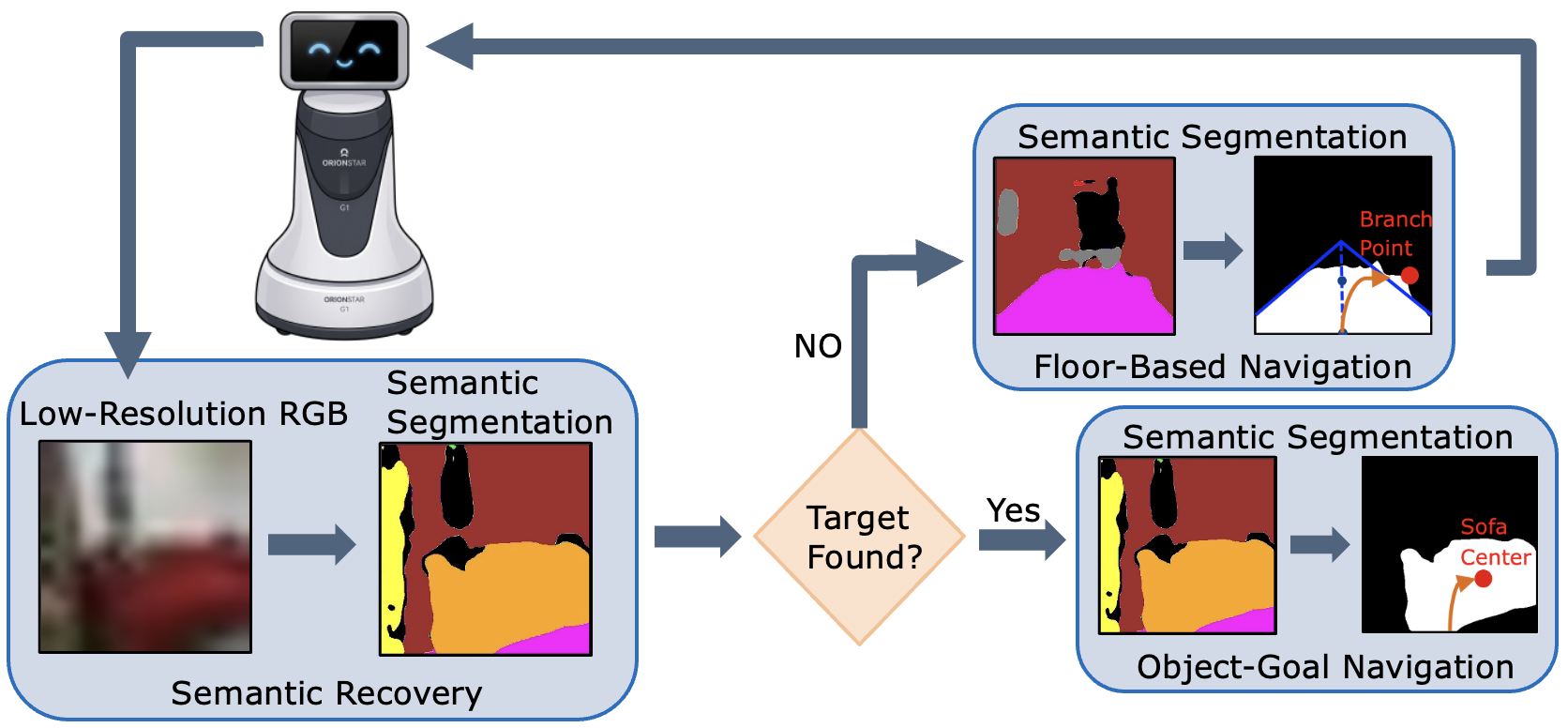}
  \caption{Overview of the adopted robot navigation pipeline. 
  The robot’s onboard camera captures an RGB image, which is first processed by a \textit{Semantic Recovery} module to generate a semantic segmentation map for downstream navigation. 
  Given a target object, the system then operates in two navigation modes. 
  If the target has not yet been detected, \textit{Floor-Based Navigation} selects the next waypoint for exploration. 
  Once the target is found, the system switches to \textit{Object-Goal Navigation} to approach the target instance.
  }
  \label{fig:system}
  \vspace{-0.6cm}
\end{figure}

\section{Study 2: User Privacy Preferences for RGB Resolution in Robot Visual Navigation}
Building on Study~1, Study 2 is designed to investigate how user privacy preferences for RGB resolution evolve during robot navigation in privacy-sensitive environments.

\subsection{RGB-to-Semantic Object-Goal Navigation System}\label{sec:system}

Study 1 revealed that users perceived semantic segmentation as a privacy-preserving visual representation and preferred privacy preservation applied at capture time through low-resolution RGB sensing. 
Guided by these findings, we instantiate our navigation setting in semantic object-goal navigation, a representative visual navigation task in which a robot is required to find and approach a target object such as a chair, desk, monitor, or sofa. 
We adopt a privacy-preserving RGB-to-semantic object-goal navigation system that supports RGB inputs at different resolutions~\citep{huang2_2025arxiv}. 
In this system, RGB observations are transformed into semantic segmentation outputs for downstream navigation.

As shown in Fig.~\ref{fig:system}, the adopted system consists of three navigation modules: Semantic Recovery, Floor-Based Navigation, and Object-Goal Navigation. 
At the beginning of navigation, the \textit{Semantic Recovery} module converts RGB images into semantic segmentation outputs for downstream navigation. 
Given a target object, the robot first uses this recovered semantic segmentation mask to search for the target. 
If the target object is not found, the \textit{Floor-Based Navigation} module determines the next waypoint for exploration. 
Once the target object is found, the system switches to the \textit{Object-Goal Navigation} module to approach it.

By supporting multi-resolution RGB inputs in a semantic navigation pipeline, this system allows us to investigate user privacy preferences in a real-world navigation setting. 
We implemented this system on the OrionStar GreetingBot Mini\footnote{\url{https://en.orionstar.com/mini.html}} mobile robot, and performed object-goal navigation in privacy-sensitive real-world environments using RGB inputs at different resolutions.

\subsection{Study Design} 
To investigate participants' RGB resolution preferences during robot navigation, we conducted another online questionnaire-based user study grounded in a real robot navigation setting. 
We adopted the privacy-preserving navigation system described in Sec.~\ref{sec:system} to collect 10 robot navigation sequences in privacy-sensitive environments with an onboard RGB camera. 
Each sequence was represented at four input resolutions~($384\times384$, $32\times32$, $16\times16$, and $8\times8$)~based on Study~1, yielding 40 resolution-specific videos in total. 
Specifically, $384\times384$ served as the high-resolution RGB reference, $32\times32$ and $16\times16$ reflected the main acceptable and preferable privacy-preserving thresholds identified in Study~1, respectively, and $8\times8$ represented the lowest resolution condition.

\begin{figure}[t]
    \centering

    \begin{subfigure}[t]{0.44\linewidth}
        \centering
        \includegraphics[width=\linewidth]{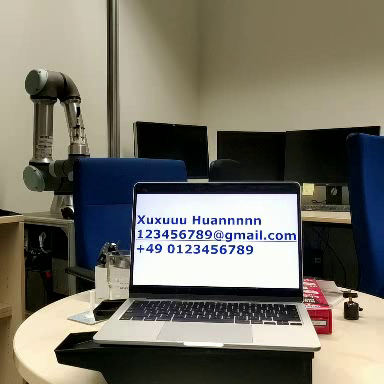}
        \caption{}
        \label{fig:nav_384}
    \end{subfigure}\hfill
    \begin{subfigure}[t]{0.45\linewidth}
        \centering
        \includegraphics[width=\linewidth]{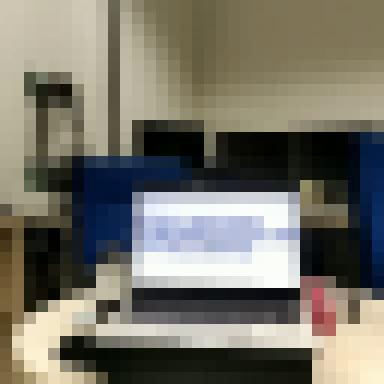}
        \caption{}
        \label{fig:nav_32}
    \end{subfigure}

    \begin{subfigure}[t]{0.45\linewidth}
        \centering
        \includegraphics[width=\linewidth]{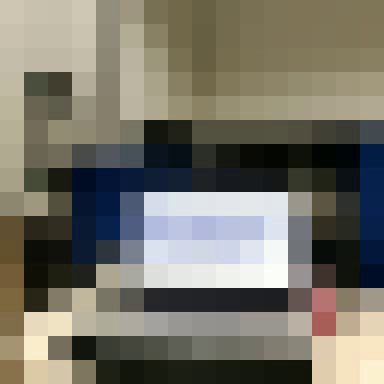}
        \caption{}
        \label{fig:nav_16}
    \end{subfigure}\hfill
    \begin{subfigure}[t]{0.45\linewidth}
        \centering
        \includegraphics[width=\linewidth]{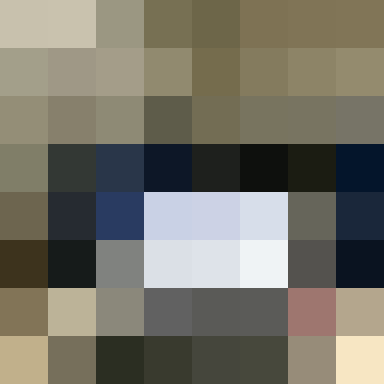}
        \caption{}
        \label{fig:nav_8}
    \end{subfigure}

    \caption{Example~(private chat) in Study~2 to examine user privacy preferences for RGB resolution at near stage. 
    The selectable options correspond to (\subref{fig:nav_384}) $384 \times 384$, (\subref{fig:nav_32}) $32 \times 32$, (\subref{fig:nav_16}) $16 \times 16$, and (\subref{fig:nav_8}) $8 \times 8$. For example, participants may judge that private information is present in (\subref{fig:nav_32}) but not recognizable, whereas no private information is perceived in (\subref{fig:nav_8}).
    }
    \label{fig:lab}
    \vspace{-0.3cm}
\end{figure}

The data cover five indoor scenes: office, lab, living room, corridor, and classroom, and includes four types of privacy-sensitive content: human face, passport, credit card, and private chat. 
The navigation targets include five objects: poster, table, chair, monitor, and object.
For each video, three frames were sampled from three stages: Beginning, Middle, and end of the navigation process~(Near). 
Following~\citep{Ferrer2013rsj}, these stages correspond to three robot-to-privacy distance ranges: Beginning~($> 3\,\mathrm{m}$), Middle~($0.9\text{--}1.2\,\mathrm{m}$), and Near~($0.3\text{--}0.45\,\mathrm{m}$).
At each stage, participants viewed a set of RGB images of the same scene at four different resolutions.
Adopting a comparative presentation strategy similar to~\citep{Khamis2024mum}, we showed participants four downsampled RGB images in a $2\times2$ arrangement, as shown in Fig.~\ref{fig:lab}.
For each stage, participants answered the same two privacy-related questions by choosing a resolution threshold: \textit{``At which resolution level and below do you think the image does \textbf{NOT} contain private information?''} and \textit{``At which resolution level and below do you think the image contains private information but it is \textbf{NOT} recognizable?''}~(see the example in Fig.~\ref{fig:lab}).
For simplicity, we refer these two questions as \textit{NOT contain} question and \textit{NOT recognizable} question.

\subsection{Study Platform} 
The study questionnaire was administered online via LimeSurvey platform\footnote{\url{https://www.limesurvey.org/}} following the same recruiting procedure described in Sec.~\ref{sec:study_1}.

\subsection{Participants} A total of N=149 individuals~(73 Men, 76 Women) participated in the study. They were on average 26.8 years old~(SD = 5.04 years).

\begin{figure}[!t] 
  \centering
  \includegraphics[width=1.0\columnwidth]{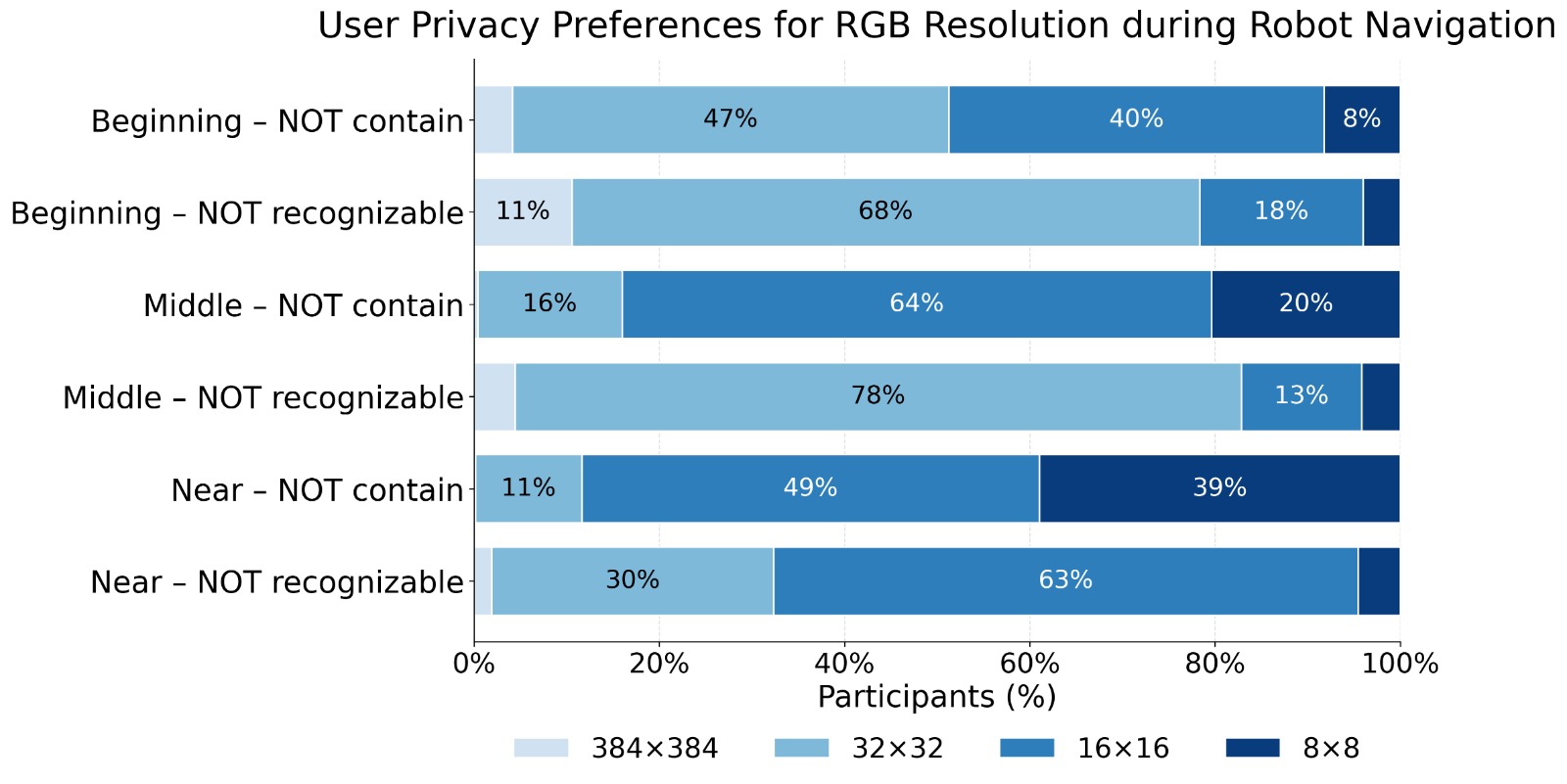}
  \caption{
  Distribution of user-selected privacy-aware RGB resolution thresholds across navigation for \textit{NOT contain} question and \textit{NOT recognizable} questions.
  The distributions suggest that participants tended to favor lower RGB resolution thresholds as robot proximity increased (N/A excluded).
}
  \label{fig:reso_nav}
  \vspace{-0.3cm}
\end{figure}

\subsection{Study Results} 
\subsubsection{User Privacy Preferences for  RGB Resolution during Robot Navigation} 
As shown in Fig.~\ref{fig:reso_nav}, under the \textit{NOT contain} question, the dominant response shifted from $32\times32$ at the Beginning stage~($> 3\,\mathrm{m}$) to $16\times16$ at the Middle stage~($0.9\text{--}1.2\,\mathrm{m}$), and then further toward $16\times16$ and $8\times8$ at the Near stage~($0.3\text{--}0.45\,\mathrm{m}$). 
Under the \textit{NOT recognizable} question, participants remained centered on $32\times32$ at both the Beginning stage and the Middle stage, but shifted toward $16\times16$ at the Near stage. 
These descriptive distributions suggest that participants tended to prefer lower RGB resolutions as the robot approached privacy-sensitive content, with this shift appearing earlier for the \textit{NOT contain} question than for the \textit{NOT recognizable} question.

To test whether navigation stage affected these ordinal threshold responses while mitigating repeated-measures dependence across participants and image items, we follow a prior work in HRI~\citep{Axelsson2023hri} and fitted Cumulative Link Mixed Models (CLMMs)~\citep{Axelsson2023hri} separately for two privacy-related questions. 

In each model, the ordinal response threshold (A--D) was treated as the dependent variable, stage (Beginning, Middle, Near) was included as a fixed effect, and random intercepts were included for participant and image item. 
Beginning stage served as the reference level, such that positive coefficients indicate shifts toward lower acceptable RGB resolutions.

For the \textit{NOT contain} question, both Middle ($\beta = 1.29$, SE $= 0.08$, $p < .001$, OR $= 3.64$) and Near ($\beta = 2.03$, SE $= 0.08$, $p < .001$, OR $= 7.59$) were associated with lower acceptable RGB resolution thresholds than Beginning. 
This suggests that, for the question of whether the image still contained private information, participants tended to tighten their preferences for RGB resolution thresholds as the robot moved from far range into middle and near range.

For the \textit{NOT recognizable} question, Middle did not differ significantly from Beginning ($\beta = 0.08$, SE $= 0.09$, $p = .367$, OR $= 1.08$), whereas Near was associated with a clear shift toward stricter thresholds ($\beta = 2.31$, SE $= 0.09$, $p < .001$, OR $= 10.08$). 
This suggests that, for the question of whether private information remained recognizable, participants' RGB resolution preferences were relatively similar between the Beginning and Middle stages, but moved toward lower resolution thresholds once the robot entered close range.

Taken together, the descriptive distributions and CLMMs results show that user privacy preferences for RGB resolution were not fixed, but varied with robot proximity during navigation. 
At farther distances, participants generally tolerated thresholds at or below moderate resolutions, whereas at closer distances they shifted toward lower resolution thresholds, which they perceived as more privacy-preserving.
This stage-dependent shift appeared earlier for the \textit{NOT contain} question and was clearly observed at the Near stage for both privacy-related questions.

\subsubsection{Objective–Subjective Alignment in Privacy Evaluation} 
To complement the subjective user study, we further conducted an objective image privacy evaluation using a model-based protocol. 
Following the protocol of Samson et al.~\citep{samson2024arxiv}, which uses a VLM-based privacy assessment model validated on the \textit{PrivBench} benchmark, we evaluated the same image conditions across the three navigation stages and four RGB resolutions.
The model outputs a \textit{privacy rate}, reflecting whether privacy-related content is detected in the image, which we interpret as the model's sensitivity to privacy-related content. 
For consistency with our user-study analysis, we report its complementary quantity, \mbox{$P$-non}, computed as $1-\textit{privacy rate}$. 
We use \mbox{$P$-non} as a model-based proxy for privacy preservation, where a larger \mbox{$P$-non} indicates that privacy-related content is less likely to be recognized by the model, and therefore corresponds to lower privacy threats under model-based recognition.

For the content-based dimension, we used the original prompt from~\citep{samson2024arxiv} to evaluate the privacy:
\textit{"Analyze the image provided. Does the attached image contain privacy information? Answer only with yes or no."}
For the recognizability-based privacy dimension, we used the prompt:
\textit{"Analyze the image provided. Does the attached image contain privacy information and if yes, is the privacy information recognizable? Answer only with yes or no."}
These two prompts correspond closely to our two user-study questions, namely whether an image contains sensitive information and whether sensitive information is recognizable. 

To enable comparison between user study results and model-based rates, we converted the human threshold responses into P-non over valid responses (excluding N/A). 
Specifically, at each stage and resolution level, the human \mbox{P-non} value represents the proportion of valid responses whose selected threshold considered that resolution sufficient to satisfy the corresponding privacy needs. 
This transformation allows the subjective threshold results and the model outputs to be compared on the same scale.

\begin{figure}[!t] 
  \centering
  \includegraphics[width=1.0\columnwidth]{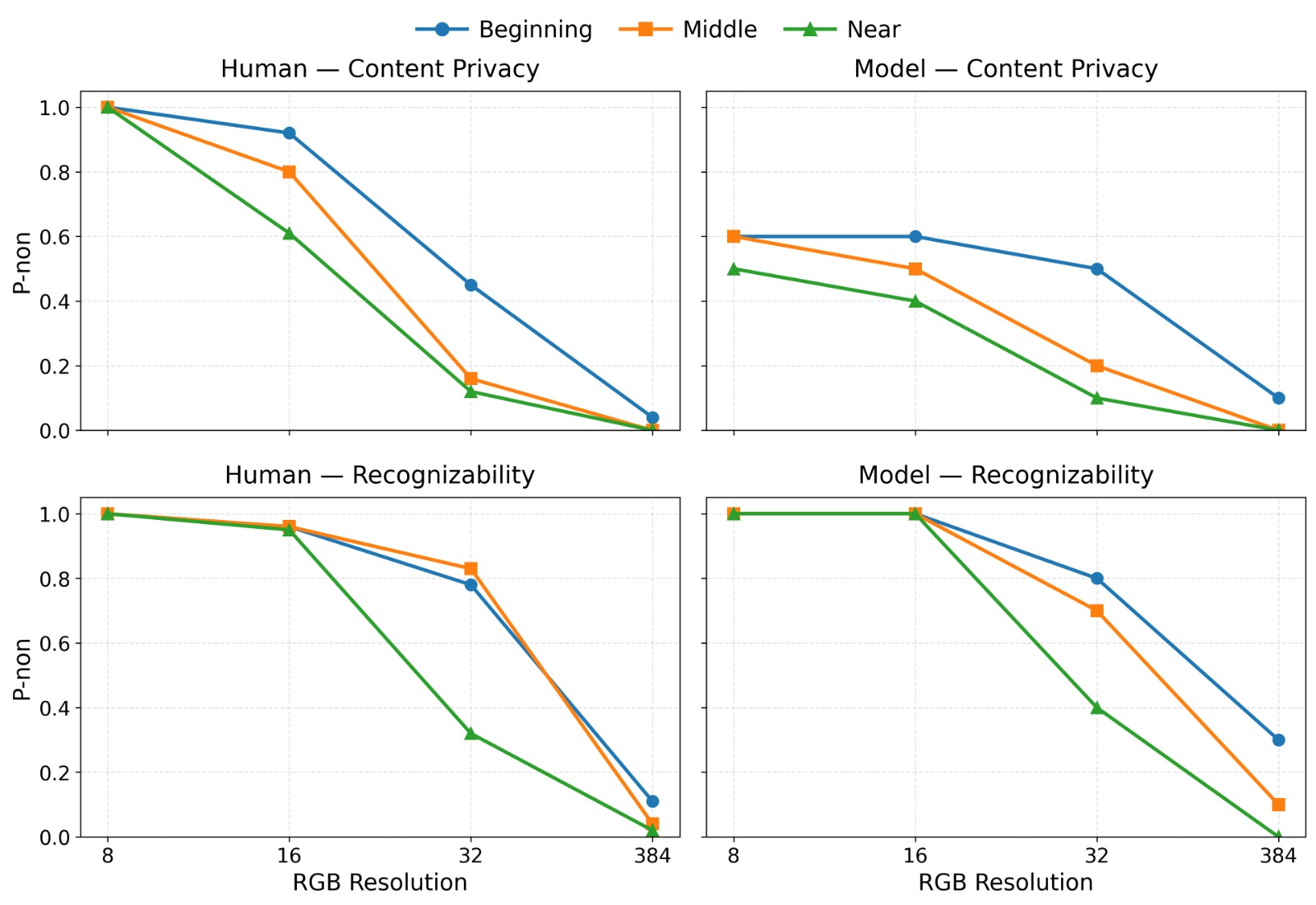}
  \caption{Comparison of subjective and objective, model-based privacy-preserving trends across RGB resolutions and robot proximity stages. 
  The left column shows human-derived P-non curves, computed from cumulative threshold responses over valid answers (excluding N/A), while the right column shows model-based P-non values obtained from the VLM-based model. 
  As can be seen, lower RGB resolutions generally yield higher \mbox{P-non} values, whereas closer robot proximity tends to reduce privacy preservation at a fixed resolution. 
  Despite differences in magnitude, both show consistent directional trends.
}
  \label{fig:vlm}
  \vspace{-0.6cm}
\end{figure}

Figure~\ref{fig:vlm} shows that the objective model exhibits the same directional trends as user preferences across both privacy dimensions. 
For a fixed RGB resolution, \mbox{P-non} decreases as robot moves from Beginning to Middle to Near, indicating increased privacy exposure at closer proximity. 
This in turn implies that lower RGB resolutions are needed at closer distances to maintain comparable privacy-preserving levels.

Under the content dimension, $32{\times}32$ provides a higher \mbox{P-non} than $384{\times}384$ at Beginning ($0.50$ vs.\ $0.10$), but this level decreases at Middle ($0.20$) and Near ($0.10$). 
By contrast, $16{\times}16$ remains associated with a higher privacy-preserving level across these stages, indicating that a level achievable at $32{\times}32$ in the Beginning stage shifts to around $16{\times}16$ once the robot enters middle range. 
A similar threshold relationship appears in the recognizability dimension. 
At the Beginning stage, $32\times32$ yields a relatively high \mbox{P-non} value ($0.80$), but this drops to $0.70$ at Middle and further to $0.40$ at Near stage. 
In contrast, $16\times16$ consistently yields the highest privacy-preserving level across all three stages.

Although the absolute magnitudes differ from the human-derived curves, the overall trends remain consistent. 
One possible reason is that, unlike human viewers, the model can still extract coarse contours and rough semantic information even from blurred images~\cite{huang2_2025arxiv}, which may lead to lower overall \mbox{P-non} values while preserving the same directional pattern across resolutions and proximity stages.

Overall, the model-based evaluation indicates not only that lower RGB resolutions are more privacy-preserving in general, but also that the privacy-preserving threshold shifts downward as robot proximity increases.
This model-based trend aligns with user preferences, as participants likewise chose lower acceptable RGB resolution thresholds when the robot moved from far to near range.

\subsubsection{Summary}
Study~2 shows that user privacy preferences for RGB resolution vary with robot proximity during navigation. 
Participants preferred progressively lower resolutions as the robot moved closer to privacy-sensitive content, with higher privacy requirements leading to earlier and stronger shifts toward lower resolutions. 
These results were supported by both the CLMMs analysis and model-based evaluation.

\section{Design Implication: User-Configurable Distance-to-Resolution Privacy Policy} 

Taken together, the findings from Study~1 and Study~2 suggest that privacy-preserving robot visual navigation should not rely on a single global RGB resolution setting, but instead be implemented through a \textit{user-configurable distance-to-resolution privacy policy}.
Study~1 shows that user privacy preferences differ across visual modalities and RGB resolution levels, while Study~2 shows that these preferences further vary with robot proximity during navigation.
More specifically, the results indicate a distance-dependent shift in acceptable RGB resolution thresholds: privacy-preserving levels achievable with moderate resolutions at farther distances require lower resolutions as the robot moves into middle or near range. 

On the other hand, while lower RGB resolutions are generally more privacy-preserving, semantic navigation still requires sufficient visual detail for reliable target recognition and scene understanding~\cite{huang2_2025arxiv, Florarxiv, Chaplotneurips}.
Therefore, rather than applying the lowest resolution uniformly, a distance-aware policy can adapt RGB resolution to balance privacy preservation with navigation quality.

Thus, the RGB input resolution should be selected as a function of robot proximity and the user’s desired privacy level. 
Users with stronger privacy sensitivity may prefer the system to switch to lower resolutions already at the middle range, whereas users with less stringent requirements may accept higher resolutions until the close range. 
This design, therefore, enables personalized privacy preservation by adapting the selected RGB resolution to robot proximity during navigation.

\textit{Discussion \& Limitation:} 
While Study~1 indicates that modalities such as depth and semantic segmentation are generally perceived as more privacy-preserving than RGB, our system-level instantiation focuses on semantic object-goal navigation based on multi-resolution RGB inputs. 
Extending user-centered privacy control to depth- or RGB-D-based navigation remains an important direction for future work, particularly given that the privacy-preserving properties of depth may vary with sensor fidelity and the amount of geometric detail retained. 
Although the static images used in Study~2 were sampled from real robot navigation sequences, they cannot fully capture users’ experiences during dynamic interaction; future studies should therefore evaluate privacy preferences through in-person interaction with the robot.

Participants in both studies were predominantly young adults, which may limit the generalizability of our findings. Future work should include more diverse populations across age, cultural background, technological familiarity, and privacy sensitivity. The VLM-based evaluation may also be affected by hallucinations and cannot generalize beyond the privacy categories represented in its training data.

\section{Conclusion}
In this work, we investigated how user privacy preferences can inform the design of privacy-preserving visual perception for robot navigation. 
Through two user studies, we showed that users generally prefer privacy-preserving visual abstractions and low-resolution RGB inputs, and that their privacy preferences for RGB resolution vary with robot proximity during navigation. 
Based on these findings, we derived a user-configurable distance-to-resolution privacy policy and showed how it can be instantiated in an existing multi-resolution RGB semantic object-goal navigation pipeline. 
Our results provide a practical step toward operationalizing user privacy preferences in navigation-oriented robot visual perception.
Future work will extend user-centered privacy-preserving navigation to RGB-D, while validating personalized and real-time adaptive privacy policies across more diverse user populations.

\section{Acknowledgements}
Generative AI, namely GPT-5.4, was used for text proofreading, and Gemini 3 Pro for the generation of Fig.~\ref{fig:teaser}.

\bibliographystyle{IEEEtranSN}
\footnotesize
\balance
\bibliography{roman2026}

\end{document}